\documentclass[11pt]{article}

\usepackage[final]{acl}

\usepackage{times}
\usepackage{latexsym}
\usepackage{amsmath,amssymb,amsfonts}
\usepackage{algorithmic}
\usepackage{graphicx}
\usepackage[T1]{fontenc}

\usepackage[utf8]{inputenc}

\usepackage{microtype}

\usepackage{inconsolata}

\usepackage{graphicx}

\title{ToneCL: Contrastive Learning for Few-Shot Syllable-Level Tone Classification}

\author{
  \textbf{Qisheng Liao}\qquad
  \textbf{Youngah Do}
  \\
  Department of Linguistics\\
  The University of Hong Kong\\
  \small{
     liaoqs@connect.hku.hk \qquad youngah@hku.hk
  }
  }

\begin{document}
\maketitle
\begin{abstract}
Tone languages constitute over 50-70\% of the world's languages, but the vast majority are low-resource, lacking the large transcribed corpora needed for automatic tone classification. Existing datasets are typically collected at the sentence level, whereas field linguists require fine-grained syllable-level annotations. We propose ToneCL, a lightweight contrastive learning framework for few-shot syllable-level tone classification. We simulate low-resource conditions on Mandarin and Vietnamese, limiting labeled data to tens of examples per tone class. ToneCL is pretrained on unlabeled speech with augmentations that preserve tonal identity, then fine-tuned on few-shot examples. Experiments show our method consistently outperforms baselines, achieving 91.6\% on six-speaker Mandarin at 10 shots. Cross-lingual transfer is also effective: pretraining on Vietnamese and fine-tuning on Mandarin reaches 91.0\% accuracy at 10 shots. Ablation confirms that frequency band rejection is the most critical augmentation.
\end{abstract}


\section{Introduction}

Recent years have witnessed remarkable progress in natural language processing, driven largely by large language models (LLMs). These models achieve state-of-the-art results on tasks such as machine translation, automatic speech recognition, and dialogue systems. However, this rapid advancement has also led to a growing imbalance: high-resource languages and large-scale tasks dominate, while linguistically meaningful problems in low-resource settings remain overlooked. Among the most underserved is tone classification for low-resource tonal languages—a task that is both theoretically important for phonological research and practically essential for language documentation, yet has received little attention in the age of LLMs.

\begin{figure}
\centering
\includegraphics[width=0.35 \textwidth]{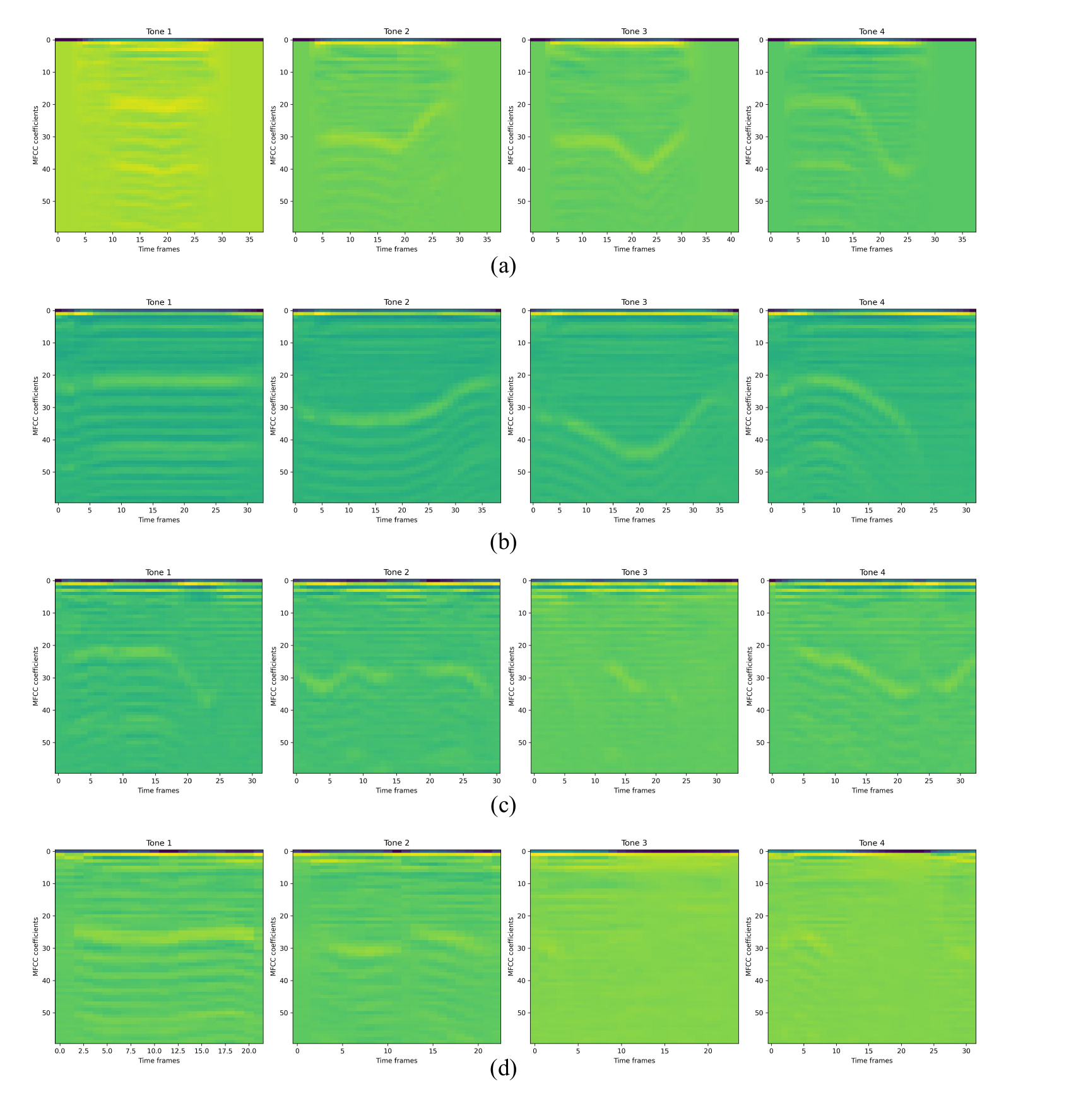}
\caption{Random MFCCs for four different Mandarin tones from (a) Google Translate text-to-speech, (b) the Tone Perfect dataset~\cite{ryu_tone_perfect}, (c) the THCHS-30 dataset~\cite{wang2015thchs}, and (d) the AISHELL-3 dataset~\cite{shi2020aishell}.}
\label{fig:1}
\end{figure}



Tone languages rely on pitch variations to distinguish lexical meaning. Over 50-70\% of the world's languages are tonal, representing more than 2,000 languages across Africa, Asia, and the Americas. Despite their prevalence, most tonal languages remain low-resource, lacking large transcribed corpora or digital tools for documentation. Most prior work on automatic tone classification has focused on a handful of high-resource languages, such as Mandarin. Low-resource tonal languages thus receive limited attention, facing severe data scarcity, lack of standardized orthographies, and restricted computational infrastructure.

A further issue concerns how existing datasets are constructed. Even for well-documented tonal languages, most speech datasets are collected at the sentence or utterance level. While this suits automatic speech recognition or utterance-level tasks, it does not support fine-grained tone analysis. In tonal languages, however, tone is often tied to individual syllables. Each syllable carries a specific tone that distinguishes it from other syllables with identical segmental content. 


To illustrate this situation, we present random MFCCs for four Mandarin tones from different datasets in Figure \ref{fig:1}. For syllable-level data, such as (a) Google Translate and (b) Tone Perfect~\cite{ryu_tone_perfect}, clear differences corresponding to each tone are readily observable. In contrast, for (c) THCHS-30~\cite{wang2015thchs} and (d) AISHELL-3~\cite{shi2020aishell}, both widely used in previous work, we applied an aligner to automatically segment the audio from sentence level to syllable level. The resulting MFCCs show weak and noisy patterns at first glance. This reveals a critical gap: the acoustic pattern of an isolated syllable suitable for language documentation differs fundamentally from that found in conversational speech. Furthermore, given the substantial difference in acoustic features, a model trained on one type of data is unlikely to generalize to out-of-domain datasets.

This mismatch poses a practical challenge for field linguists. Most available data comes in the form of continuous speech, where tonal cues are obscured by coarticulation. Existing work reports sentence-level tone error rates or frame-level accuracy—metrics of limited utility to linguists who need reliable tone annotations for individual syllables from just a handful of speakers. Yet this specific use case has received almost no attention in computational tonology.

Beyond the dataset and evaluation mismatch, there exists a broader practical gap between NLP research and the needs of field linguists. First, models must be lightweight and easy to train. Large speech foundation models like wav2vec 2.0 \cite{baevski2020wav2vec}, while powerful, require substantial GPU resources and training time. Second, these models demand large amounts of data, which are impossible to collect during a short field trip, especially for undocumented low-resource languages. Third, there is a fundamental architectural mismatch: wav2vec 2.0 and similar models are designed for several seconds of continuous speech, relying on temporal context to build representations. A single tone-bearing syllable, however, is typically under one second. On such short inputs, these models fail to leverage their contextual encoders and produce inconsistent outputs, a finding we confirm empirically in Section~\ref{mainexp}. Fourth, even the choice of input representation remains unsettled. Should we use raw waveforms, spectrograms, or MFCCs? Should we rely solely on F0, or incorporate other acoustic features? These open questions pose practical barriers for linguists who wish to adopt machine learning tools.

In addition to these practical concerns, cross-lingual transfer presents another critical challenge. Many low-resource tonal languages have better-documented relatives, but direct transfer is nontrivial due to differences in tone inventories, phonetic realization, and phonation. A method that adapts representations from a high-resource to a low-resource language with minimal labeled data would be highly valuable, yet this remains largely unexplored.

To address these challenges, we propose ToneCL, a lightweight contrastive learning framework for syllable-level tone classification. Since no public syllable-level dataset exists for truly low-resource languages, we simulate such conditions using Mandarin and Vietnamese by constraining labeled data to tens of examples per tone. ToneCL pretrains on unlabeled speech with carefully designed augmentations that preserve tonal identity, then fine-tunes on few-shot examples, achieving consistent gains in multi-speaker few-shot settings. It reaches 91.6\% accuracy at 10 shots on six-speaker Mandarin and demonstrates effective cross-lingual transfer. This suggests that for typologically similar languages, unlabeled data from a related high-resource language can serve as a practical substitute when target data is scarce.

Our main contributions are:
\begin{itemize}
\item First few-shot syllable-level tone classification framework. To our knowledge, this is the first work for this task under low-resource constraints, requiring only tens of labeled examples.
\item Cross-lingual transfer potential. We show that pretraining on one tonal language can improve classification in another, suggesting potential for leveraging related high-resource languages.
\item Comprehensive ablation study. Our experiments validate key design choices, including which augmentation strategies best preserve tonal identity and improve performance.
\end{itemize}

\section{Related Work}

\subsection{Tone Classification}

 A substantial body of prior work has addressed Mandarin tone classification using various methods, including convolutional neural networks (CNNs) \cite{chen2016tone, gao2019tonenet}, RNNs \cite{huang2021encoder}, Transformers \cite{liu2024learning}, wav2vec 2.0 \cite{yuan2021automatic, yuan2023improved}, and even GANs \cite{schenck2025unsupervised}. Some researchers have proposed specialized frameworks such as Tone2vec \cite{yang2024automated}. While these methods report strong performance, they typically rely on training datasets that are not limited in size, making them difficult to extend to low-resource tonal languages. Furthermore, some of these approaches operate at the sentence level using CTC models and report tone error rates, a metric that is not directly useful for linguistic documentation tasks that require syllable-level annotations.

Beyond Mandarin, several studies have investigated tone classification for low-resource tonal languages. For example, recent work on Yoruba applied wav2vec 2.0 to tone classification \cite{10.1145/3690384}. Similarly, \cite{gogoi2021learning} introduced a CNN-based method to classify tones in Mizo, a Sino-Tibetan language spoken in India, using F0 contours. However, these methods operate at the sentence level rather than on individual syllables, and their reported accuracy remains insufficient for practical documentation tasks. More critically, sentence-level predictions cannot be easily decomposed into syllable-level tone annotations, which are precisely what field linguists need for tasks such as dictionary building and tone pattern verification.

\subsection{Contrastive Learning}
Contrastive learning has emerged as a powerful paradigm for self-supervised representation learning. The core idea is to learn representations by pulling semantically similar examples close together in the embedding space while pushing dissimilar examples apart. This approach was first widely adopted in computer vision, where frameworks such as SimCLR \cite{chen2020simple} and MoCo \cite{he2020momentum} demonstrated that contrastive pretraining on unlabeled data could match or exceed supervised learning performance.

The success of contrastive learning has since extended to natural language processing. Models such as SimCSE \cite{gao2021simcse} and ConSERT \cite{yan2021consert} apply contrastive objectives to learn sentence representations from unlabeled text through simple dropout-based or augmentation-based positive pair construction. In speech processing, contrastive methods have been incorporated into frameworks like wav2vec 2.0 \cite{baevski2020wav2vec} and HuBERT \cite{hsu2021hubert}, where the objective is to distinguish a masked target frame from distractors. However, these speech models are typically large and require substantial data, limiting their applicability to low-resource settings.

\begin{figure*}
\centering
\includegraphics[width=0.83 \textwidth]{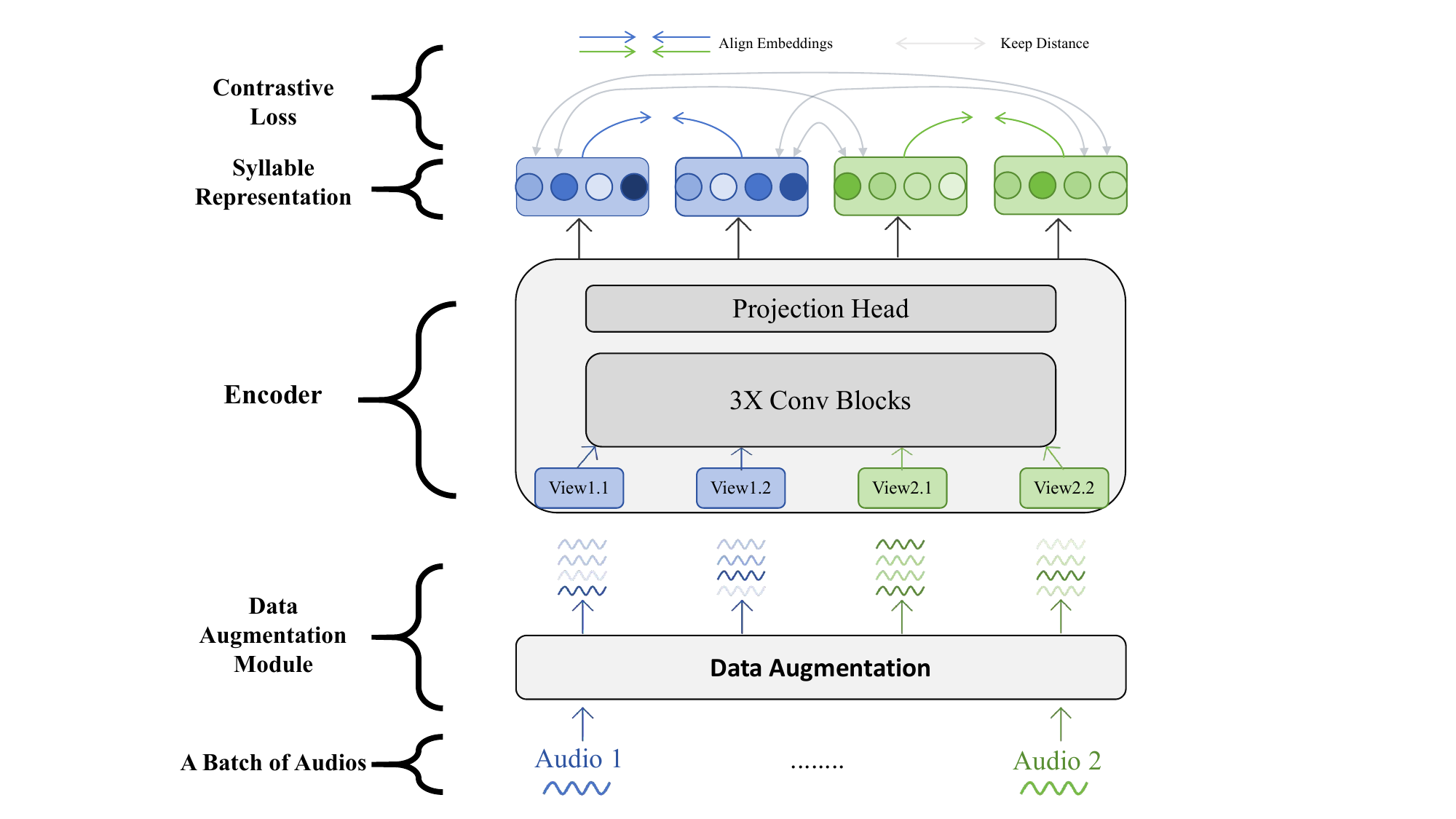}
\caption{The general framework of our proposed approach.}
\label{fig:2}
\end{figure*}

\section{Method}

\subsection{Preliminaries: Contrastive Learning}

Contrastive learning learns representations by pulling semantically similar examples together and pushing dissimilar examples apart in the embedding space. Given a batch of $N$ examples, for each example $x$, we create a positive pair by applying two different augmentations to the same source, yielding $x_i$ and $x_j$. The remaining $2(N-1)$ augmented examples in the batch serve as negative pairs. The standard InfoNCE loss \cite{chen2020simple} is defined as:
\begin{equation}\label{eq:1}
\mathcal{L} = -\frac{1}{N}\sum_{i=1}^{N}\log
\frac{\exp(\operatorname{sim}(z_i,z_j)/\tau)}
{\sum_{\substack{k=1\\k\neq i}}^{2N}\exp(\operatorname{sim}(z_i,z_k)/\tau)}
\end{equation}
where $j$ indexes the positive view paired with anchor $i$, $z_i=f_\theta(x_i)$ is the normalized embedding produced by encoder $f_\theta$, $\operatorname{sim}(\cdot,\cdot)$ denotes cosine similarity, and $\tau$ is a temperature hyperparameter. The objective encourages the model to learn representations that are invariant to augmentations while remaining discriminative across different instances.

\subsection{General Framework}

Our approach is mainly inspired by SimCLR \cite{chen2020simple}. As shown in Figure \ref{fig:2}, there are three major components in our framework:

\begin{itemize}
\item Data augmentation module. This module generates two correlated views for each input audio sample through a random composition of transformations applied in the time and spectrogram domains. No original view is preserved. The augmentations are designed to preserve tonal identity while introducing controlled diversity.

\item Lightweight encoder. We use a small encoder consisting of three convolutional layers followed by a projection head. The encoder computes syllable-level representations for each input audio.

\item Contrastive loss. The loss function aligns embeddings from the two views of the same source syllable while pushing representations from different syllables apart.
\end{itemize}

\subsection{Self-Supervised Pretraining for Tone Classification}

While contrastive learning has been successful in computer vision, its application to tone classification faces a unique challenge: designing effective augmentations that preserve tonal identity while generating diverse positive pairs. In vision, augmentations such as random cropping, flipping, and color jitter are natural and label-preserving. In natural language processing, methods like SimCSE \cite{gao2021simcse} use dropout as a minimal augmentation or leverage semantic similarity through paraphrase generation. However, neither approach translates directly to tone classification. Semantic transformations are ill-suited for acoustic tone data, and standard audio augmentations such as random cropping risk destroying the very tonal information we aim to capture.

To address this, we design a set of augmentations specifically for syllable-level tone classification. Our goal is to create diverse positive pairs that preserve the tonal contour, while introducing controlled variation to improve robustness. Before we augment the data, we first segment each unlabeled utterance to the first second of audio. This serves two purposes: (1) it reduces computational overhead, and (2) it provides a fixed-length input that is sufficient to capture tonal information for most syllables.

\subsection{Augmentation strategies}
All augmentations are applied to unlabeled speech data during pretraining. We apply the following transformations to create positive pairs:

\paragraph{Time stretching.} We randomly adjust the audio playback speed by a factor between 0.9 and 1.1 using phase vocoder-based resampling. This alters temporal dynamics while preserving pitch contour, encouraging the model to learn tempo-invariant tonal representations.

\paragraph{Gaussian noise injection.} We add Gaussian noise with amplitude randomly sampled from 0.001 to 0.005. This introduces subtle background noise that improves model robustness against imperfect recording conditions commonly encountered in field documentation settings, without overwhelming the tonal signal.

\paragraph{Volume perturbation.} We randomly adjust the gain by scaling the amplitude between -6 dB and +6 dB. This makes the model invariant to variations in recording volume, which often differ across sessions, speakers, or recording devices.

\paragraph{Frequency band rejection.} We apply a bandpass filter with a center frequency randomly chosen between 200 Hz and 4000 Hz using a fixed bandwidth. Lower center frequencies produce muffled audio dominated by low-frequency components, while higher center frequencies yield brighter but still band-limited output. This random variation forces the model to learn robust representations that are not tied to any specific frequency range, while still preserving the fundamental frequency (F0) region most informative for tone identification and discarding very low-frequency noise and high-frequency artifacts.

\paragraph{Low-frequency masking.}
We randomly mask a small number of low-frequency bins (1–2 consecutive bins) within the first 5 bins. This region corresponds to the fundamental frequency (F0) range where tonal information is primarily encoded. Masking here is applied sparingly to avoid destroying critical tonal cues.


\paragraph{Time masking.} We randomly mask 4–8 consecutive time frames in the spectrogram. This encourages the model to capture temporal patterns and transitional information rather than relying on specific short-time frames.

\paragraph{Tone contour enhancement.} We add random noise to the low-frequency region (first 5 frequency bins) of the spectrogram, scaled by 0.05. This subtly perturbs the pitch-relevant region while preserving the overall tonal contour, making the model robust to natural pitch variations across speakers and speaking styles.

All augmentations are applied stochastically with probability 0.4 during pretraining. For each unlabeled sample, we generate two correlated views by applying a random composition of the above transformations. The contrastive objective then encourages the encoder to learn representations that are invariant to these acoustically diverse but tonally faithful augmentations.

\subsection{Few-Shot Fine-Tuning}

After pretraining, we adapt the encoder to downstream tone classification using only tens of labeled syllables per target language. We discard the projection head and attach a linear classification layer.

The required number of labeled examples depends on the diversity of the data. When the dataset contains multiple speakers, we typically need approximately 10 samples per tone to achieve reasonable accuracy. For datasets with only a single speaker, 3 samples per tone are sufficient. This low requirement makes our method practical for field linguists, who often work with a small number of informants.

\section{Experiments}

\subsection{Datasets}

We use four datasets in our main experiments, covering two tonal languages: Mandarin Chinese and Vietnamese.

\paragraph{THCHS-30} \cite{wang2015thchs} is a Mandarin Chinese speech corpus containing over 30 hours of read speech from 30 speakers.

\paragraph{VIVOS} \cite{luong-vu-2016-non} is a Vietnamese speech corpus consisting of around 15 hours of read speech from 15 speakers.

\paragraph{Tone Perfect} \cite{ryu_tone_perfect} is a Mandarin Chinese dataset designed specifically for tone classification. It contains isolated syllables with clear tone labels. We use this dataset for few-shot fine-tuning and evaluation. There are 6 speakers (3 male, 3 female). We also select the first female speaker as the one-speaker subset.

\paragraph{Google Text-to-Speech Vietnamese} We generate a Vietnamese tone classification dataset using Google Translate Text-to-Speech. We collect the top 10k most commonly used Vietnamese words and generate the dataset. The generated voice is female.

\begin{table*}[t]
\small
\centering
\begin{tabular}{lccccc}
\hline
\multicolumn{1}{c}{\textbf{Model}} & \textbf{\#Params} & \multicolumn{4}{c}{\textbf{Samples Per Tone}} \\
\cline{3-6}
& & 1 & 3 & 5 & 10 \\
\hline
\multicolumn{6}{c}{\textbf{Tone Perfect Mandarin (One Speaker)}} \\
\hline
CNN  & 0.5M & 71.6 (10.7) & 89.1 (7.5) & 91.6 (5.3) & 96.3 (3.0) \\
CNN-DAE \cite{chen2016tone} & 0.2M & 55.1 (1.8) & 85.4 (1.5) & 89.3 (1.5) & 92.8 (1.4) \\
ToneNet \cite{gao2019tonenet} & 1.6M & 33.4 (2.2) & 84.3 (2.1) & 86.3 (2.1) & 92.0 (1.9) \\
HuBERT \cite{hsu2021hubert} & 95M & 34.0 (2.2) & 67.3 (4.1) & 87.5 (2.2) & 89.5 (2.3) \\
wav2vec2-xlsr-300m \cite{babu2022xls} & 300M & 29.1 (4.1) & 34.4 (4.6) & 50.2 (16.9) & 91.1 (7.1) \\
ToneCL-ZH (Mandarin) & 0.5M & \textbf{80.2} (10.2) & \textbf{94.1} (3.5) & \textbf{95.1} (3.0) & \textbf{98.3} (0.9) \\
ToneCL-VN (Vietnamese) & 0.5M & 79.8 (8.6) & 93.3 (4.5) & 94.3 (3.1) & 97.1 (0.8) \\
\hline
\multicolumn{6}{c}{\textbf{Tone Perfect Mandarin (Six Speakers)}} \\
\hline
CNN  & 0.5M & 50.6 (2.8) & 68.6 (1.8) & 77.1 (1.5) & 87.8 (2.1) \\
CNN-DAE \cite{chen2016tone} & 0.2M & 34.6 (1.2) & 57.5 (1.7) & 65.4 (1.6) & 72.8 (0.7) \\
ToneNet \cite{gao2019tonenet} & 1.6M & 34.2 (2.0) & 52.6 (2.2) & 50.4 (2.0) & 60.8 (1.8) \\
HuBERT \cite{hsu2021hubert} & 95M & 38.8 (1.9) & 48.2 (2.6) & 61.1 (5.0) & 85.7 (3.2) \\
wav2vec2-xlsr-300m \cite{babu2022xls} & 300M & 26.6 (2.0) & 38.3 (10.4) & 40.0 (6.0) & 63.1 (20.2) \\
ToneCL-ZH (Mandarin) & 0.5M & \textbf{55.1} (1.0) & \textbf{76.4} (1.2) & \textbf{81.4} (1.2) & \textbf{91.6} (0.7) \\
ToneCL-VN (Vietnamese) & 0.5M & 54.2 (1.8) & 75.4 (0.7) & 80.4 (1.2) & 91.0 (0.7) \\
\hline
\multicolumn{6}{c}{\textbf{Google TTS Vietnamese (One Speaker)}} \\
\hline
CNN  & 0.5M & 42.4 (2.5) & 62.3 (1.8) & 81.5 (2.6) & 89.4 (1.1) \\
CNN-DAE \cite{chen2016tone} & 0.2M & 55.5 (2.0) & 69.3 (0.9) & 86.2 (0.5) & 92.5 (0.4) \\
ToneNet \cite{gao2019tonenet} & 1.6M & 40.2 (1.3) & 64.3 (1.2) & 78.9 (0.6) & 88.3 (0.6) \\
HuBERT \cite{hsu2021hubert} & 95M & 32.0 (1.6) & 44.0 (2.6) & 57.6 (9.7) & 70.7 (9.6) \\
wav2vec2-xlsr-300m \cite{babu2022xls} & 300M & 15.3 (8.1) & 20.7 (4.7) & 28.5 (10.1) & 35.2 (8.8) \\
ToneCL-ZH (Mandarin) & 0.5M & 63.2 (3.0) & 80.4 (3.0) & 88.5 (1.3) & 92.8 (0.4) \\
ToneCL-VN (Vietnamese) & 0.5M & \textbf{68.1} (1.8) & \textbf{81.2} (2.8) & \textbf{88.8} (2.0) & \textbf{93.2} (0.5) \\
\hline
\end{tabular}
\caption{Few-shot tone classification accuracy (\%), with standard deviations in parentheses.}
\label{tbl:1}
\end{table*}

\subsection{Experimental Setup}
\label{exp_sec}

\paragraph{Model Architecture.}
The encoder consists of three convolutional blocks (32, 48, and 120 channels) with $2\times2$ kernels, batch normalization, ReLU, and $2\times2$ max-pooling. Adaptive average pooling ($3\times5$) produces a 1,800-dimensional feature vector, which is projected via a two-layer MLP (256 hidden units, ReLU) to a 128-dimensional L2-normalized embedding. The model has approximately 524K trainable parameters.

\paragraph{Training Details.}
We use the InfoNCE loss with temperature 0.07. Pretraining runs for 500 epochs with a learning rate of $5\times10^{-4}$ and batch size of 16 on an NVIDIA RTX 4090 GPU. Fine-tuning uses $5\times10^{-5}$ learning rate for 200 epochs.

\paragraph{Evaluation Protocol.}
We reserve 2,000 samples for validation and 1,000 for testing (1000 and 500 for each Tone Perfect one speaker split). Few-shot training sets (1, 3, 5, and 10 shots per tone) are sampled from the remaining data. All results are reported as the mean and standard deviation over five random seeds.

The implementation of the main experiments is provided in the supplementary software archive.

\paragraph{Baseline Models.}
We compare against several baselines. CNN is our lightweight encoder trained from scratch. CNN-DAE \cite{chen2016tone} uses a denoising autoencoder for pretraining. ToneNet \cite{gao2019tonenet} is a CNN trained from scratch on raw spectrograms. HuBERT \cite{hsu2021hubert} (95M parameters) and wav2vec2-xlsr-300m \cite{babu2022xls} (300M parameters) are large-scale self-supervised speech models; we fine-tune them on our few-shot data without architectural changes. Our proposed ToneCL uses contrastive pretraining with tonally-preserving augmentations.

\subsection{Main Experiment}
\label{mainexp}
We pretrain ToneCL on unlabeled sentence-level data from THCHS-30 \cite{wang2015thchs} (Mandarin) or VIVOS \cite{luong-vu-2016-non} (Vietnamese), retaining only the first second of each audio sample. ToneCL-ZH and ToneCL-VN denote versions pretrained on Mandarin and Vietnamese, respectively.

We then evaluate in-language and cross-lingual transfer on Tone Perfect (Mandarin) \cite{ryu_tone_perfect} and Google TTS Vietnamese. For Tone Perfect, we report both per-speaker performance (averaged over six individual speakers) and the combined six-speaker setting. Results are reported in Table \ref{tbl:1}. 

First, contrastive pretraining consistently improves performance over the baseline across all settings and datasets. For the six-speaker split of Tone Perfect, the baseline achieves 50.6\% accuracy in the 1-shot setting. After pretraining with Mandarin data, accuracy improves to 55.1\% (an increase of 4.5\%). At higher shot counts, the improvements are consistent: from 77.1\% to 81.4\% at 5 shots, and from 87.8\% to 91.6\% at 10 shots, making the model practically usable for documentation tasks.

Second, cross-lingual pretraining shows strong potential. For the six-speaker Mandarin task, Vietnamese pretraining achieves 91.0\% accuracy at 10 shots, only 0.6\% behind Mandarin pretraining (91.6\%). For the one-speaker Vietnamese task, Mandarin pretraining reaches 92.8\%, slightly below Vietnamese pretraining (93.2\%) but still highly competitive. These results suggest that, for language pairs like Mandarin and Vietnamese, tonal features share sufficient common structure to enable effective cross-lingual transfer. While transfer may not succeed for all language pairs, this demonstrates potential for low-resource documentation scenarios where unlabeled data from a related high-resource language may be available even when target language data is scarce.

Third, task difficulty varies substantially across datasets. The six-speaker Mandarin dataset is the most challenging, with the baseline at 10 shots reaching 87.8\%. The one-speaker Mandarin dataset achieves 96.3\% at 10 shots due to the absence of speaker variation. The Vietnamese dataset falls in between, with the baseline at 89.4\% at 10 shots. Despite these differences, pretraining delivers consistent improvements across all conditions.

Fourth, the choice of pretraining language matters little with sufficient target labels. The gap between in-language and cross-lingual pretraining narrows as more labeled data becomes available, dropping to within 1\% at 10 shots, suggesting that with sufficient target labels, the source language matters little.

The Vietnamese TTS dataset provides a controlled single-speaker benchmark, while Tone Perfect evaluates performance on human speech. ToneCL reaches 93.2\% on Vietnamese TTS and 98.3\% on one-speaker Mandarin at 10 shots.

\subsection{Large Pretrained Models Struggle in Few-Shot Settings}

We further evaluate two large-scale self-supervised speech models: HuBERT \cite{hsu2021hubert} (95M parameters) and wav2vec2-xlsr-300m \cite{babu2022xls} (300M parameters). Both models are designed for sentence-level inputs of several seconds and rely on rich contextual representations. Table \ref{tbl:1} shows that they perform poorly under our few-shot, syllable-level conditions.

First, both models exhibit high variance across random seeds, with standard deviations substantially larger than our ToneCL (e.g., wav2vec on one-speaker Mandarin at 10 shots: 91.1\% $\pm$ 7.1\%; HuBERT on Vietnamese TTS at 10 shots: 70.7\% $\pm$ 9.6\%). This instability makes them unreliable for low-resource documentation. Second, they fail to converge in many low-shot settings: wav2vec achieves only 15.3\% at 1 shot on Vietnamese TTS, barely above random. Third, even at higher shot counts, their performance lags behind ToneCL on the challenging six-speaker Mandarin task (e.g., 85.7\% for HuBERT vs. 91.6\% for ToneCL at 10 shots).

We attribute this to an architectural mismatch: HuBERT and wav2vec are pretrained on continuous speech with a large receptive field, expecting several seconds of context. Our syllable-level inputs (under 1 second) provide insufficient context for these models to leverage their pretrained representations, leading to unstable fine-tuning. This confirms that large speech foundation models are ill-suited for syllable-level few-shot tone classification, reinforcing the need for lightweight, task-specific frameworks like ToneCL.

\subsection{Comparison with Prior Work for Tone Classification}

We compare against two prior CNN-based approaches: CNN-DAE \cite{chen2016tone} (pretrained with a denoising autoencoder) and ToneNet \cite{gao2019tonenet} (trained from scratch). Although ToneNet reportedly outperforms CNN-DAE on standard benchmarks, it struggles under our few-shot constraints, reaching only 60.8\% on six-speaker Mandarin at 10 shots versus CNN-DAE's 72.8\%. CNN-DAE benefits from pretraining, achieving competitive results on one-speaker Mandarin (92.8\% at 10 shots, close to our 98.3\%). However, in challenging settings, our method shows clear advantages: on six-speaker Mandarin at 1 shot, CNN-DAE achieves 34.6\% versus our 55.1\%; at 3 shots, 57.5\% versus 76.4\%. Our method outperforms ToneNet by 20.9\% at 1 shot and 23.8\% at 3 shots. This demonstrates that our contrastive framework is substantially more effective under extreme data scarcity.

\begin{table}[t]
\small
\centering
\begin{tabular}{lcccc}
\hline
\textbf{Pretrain Samples} & \multicolumn{4}{c}{\textbf{Samples Per Tone}} \\
\cline{2-5}
& 1 & 3 & 5 & 10 \\
\hline
50 & 53.0 & 75.7 & 79.4 & 90.2 \\
100 & 55.1 & 76.4 & 81.4 & 91.6 \\
500 & 57.8 & 78.1 & 82.8 & 92.1 \\
1,000 & 58.9 & 80.1 & 84.3 & 92.5 \\
\hline
\end{tabular}
\caption{Few-shot tone classification accuracy (\%) using different amounts of unlabeled pretraining data.}
\label{tbl:pretrain_config}
\end{table}

\begin{table*}[t]
\small
\centering

\begin{tabular}{lcccc}
\hline
\textbf{Configuration} & \multicolumn{4}{c}{\textbf{Samples Per Tone}} \\
\cline{2-5}
& 1 & 3 & 5 & 10 \\
\hline
CNN baseline & 50.6 & 68.6 & 77.1 & 87.8 \\
\hline
\multicolumn{5}{c}{\textit{Full augmentation}} \\
Stochastic, $p{=}0.4$ & \textbf{55.1}  & \textbf{76.4}  & \textbf{81.4} & \textbf{91.6}\\
Deterministic, $p{=}1.0$ &55.0  & 75.8 &80.3  & 90.1 \\
\hline
\multicolumn{5}{c}{\textit{Single augmentation only}} \\
Only Time stretching &52.1  & 75.8 & 78.9 & 90.0 \\
Only Gaussian noise & 51.9 & 74.7 & 79.1 & 89.1 \\
Only Volume perturbation & 50.3 & 75.2 & 76.9 & 90.0 \\
Only Frequency band rejection & 54.5 & 76.0 & 79.3 & 91.0 \\
Only Low-frequency masking & 51.5 & 73.5 & 79.3 & 89.8 \\
Only Time masking & 50.9 & 75.4 & 79.0 & 89.5 \\
Only Tone contour enhancement & 52.1 & 72.7 & 77.3 & 90.0 \\
\hline
All except Frequency band rejection &53.2  & 76.4 & 79.2 & 90.4 \\
\hline
\end{tabular}

\caption{Ablation study on augmentation strategies.}
\label{tbl:augmentation}
\end{table*}

\section{Ablation Study}

In our ablation experiments, we pretrain on THCHS-30 \cite{wang2015thchs} using 100 unlabeled samples and fine-tune on the six-speaker split of Tone Perfect \cite{ryu_tone_perfect} with 1, 3, 5, and 10 shots per tone. All results are reported as the mean over five random seeds.

\subsection{Effect of Pretraining Data Size}

We investigate how the amount of unlabeled pretraining data affects downstream few-shot performance on the six-speaker Mandarin task. Table \ref{tbl:pretrain_config} reports results using 50, 100, 500, and 1,000 pretraining samples.

Performance improves as the amount of pretraining data increases, but the gains become marginal beyond 100 samples. For example, increasing the pretraining set from 100 to 500 samples yields only a 0.5\% improvement at 10 shots. Since 100 samples correspond to roughly two minutes of audio, we use this setting as the default for our ablation studies.

\subsection{Effect of Augmentations}

We evaluate the contribution of each augmentation strategy to the final performance. Table \ref{tbl:augmentation} presents the results, with the CNN baseline (50.6\%, 68.6\%, 77.1\%, 87.8\%) and our full method (each augmentation applied with 40\% probability) shown for reference.

\paragraph{Stochastic vs. deterministic augmentation.}
In our default setting, each augmentation is applied independently with probability $p=0.4$ per sample. This stochastic strategy achieves 91.6\% at 10 shots. Applying all augmentations deterministically ($p=1.0$) yields slightly lower performance (90.1\% at 10 shots), suggesting that stochastic application provides better regularization.

\paragraph{Individual augmentations.}
Each augmentation alone improves over the CNN baseline. At 10 shots, individual augmentations achieve between 89.1\% and 91.0\%, all substantially above the baseline of 87.8\%.

\paragraph{The importance of frequency band rejection.}
Frequency band rejection stands out as the most critical augmentation. When applied alone, it achieves 91.0\% at 10 shots, the highest among single augmentations and competitive with the full stochastic method (91.6\%).

\paragraph{Removing frequency band rejection.}
When we remove frequency band rejection from our augmentation set while keeping all others, performance drops to 90.4\% at 10 shots, a decline of 1.2 percentage points from the full method. This confirms the central role of frequency band rejection.

\paragraph{Summary.}
Our full method with stochastic augmentation ($p=0.4$) achieves the best overall performance (91.6\% at 10 shots). While frequency band rejection is the most powerful single augmentation, the combination of all augmentations yields the most robust results across varying shot counts.

\section{Conclusion}

We proposed a lightweight contrastive learning framework for few-shot syllable-level tone classification, to our knowledge the first work specifically designed for this task in low-resource language documentation scenarios. Our model pretrains on unlabeled speech with tonally-preserving augmentations and fine-tunes on only tens of labeled examples. Experiments on Mandarin and Vietnamese show consistent improvements over baselines, achieving consistent gains in challenging multi-speaker few-shot settings, reaching 91.6\% accuracy at 10 shots on six-speaker Mandarin, as well as promising cross-lingual transfer between the two languages. Ablation studies validate the importance of our augmentation strategies.

We wish to highlight a broader issue that extends beyond methodology. Even for well-studied languages like Mandarin, there is a striking lack of publicly available datasets designed for syllable-level tone classification. Most existing speech corpora are collected at the sentence level, and automatically segmenting them into syllables yields noisy, unreliable annotations (Figure \ref{fig:1}). For low-resource tonal languages, the situation is even more dire. We encourage the NLP community to pay greater attention to the practical needs of field linguists and language documenters. Creating clean, syllable-level, few-shot datasets would significantly accelerate progress in this underserved area. We hope our work serves as a first step toward bridging the gap between contrastive learning research and real-world linguistic documentation.


\section*{Limitations}
We acknowledge several limitations of this work.

\paragraph{Isolated syllables vs. naturalistic speech.}
Our models are trained and evaluated on isolated or pseudo-isolated syllables (Tone Perfect and Google TTS). In naturalistic speech, tone interacts with segmental context, coarticulation, phonation, syllable structure, and prosodic position. As shown in Figure \ref{fig:1}, forced-alignment extraction of syllable-level tones from continuous speech yields noisy features that are currently unsuitable for documentation purposes. However, we acknowledge that real fieldwork involves raw, naturalistic production. Our focus on isolated syllables is a first step toward building tools for low-resource language documentation, where even basic dictionary forms are often unavailable. Extending our method to handle continuous speech with robust syllable extraction remains an important direction for future work.

\paragraph{Cross-lingual transfer.}
We demonstrate effective transfer between Mandarin and Vietnamese, but both are East Asian contour-tone languages with shared acoustic cues (F0 height and contour). The strong transfer performance may partly reflect phonetic overlap rather than full phonological generalization. We do not claim universal transferability across all tonal languages, such as African register tone languages or languages with complex tone sandhi. Our results are best interpreted as showing potential: if a high-resource language (even unlabeled data) is available for a typologically similar low-resource language, our framework may provide benefit. Whether transfer succeeds for other language pairs requires further investigation.

\paragraph{Single-syllable focus and phonological phenomena.}
Our work focuses on single tones in isolation, following the minimal unit of tone in languages like Mandarin where each character carries a citation tone. However, we acknowledge that this excludes important tonal phenomena such as tone sandhi, tonal coarticulation, and anticipatory/carryover effects, which are central to both Mandarin and Vietnamese. Thus, our findings do not directly speak to how tonal systems behave in connected speech or multi-syllable contexts. The generalizability of our approach to broader tonal phenomena remains an open question.

\paragraph{Phonation.}
Vietnamese tones involve phonation contrasts (breathy, creaky, glottalized) that are not explicitly modeled by our MFCC-based input, which primarily captures spectral envelope. While MFCCs may partially encode spectral tilt correlates of phonation, we do not claim to fully capture these cues. The perception literature suggests that phonation serves as an important secondary cue for tone perception, though F0 remains primary \cite{lu2025contribution}. Future work should incorporate dedicated voice quality features to better handle languages with phonation-based tone contrasts.

\paragraph{Comparison with prior work.}
We note that the two prior works we compared against \cite{chen2016tone,gao2019tonenet} originally reported results under conditions with abundant labeled data (hundreds or thousands of examples per tone). Their models were not designed for few-shot settings, and our reimplementation may not represent their optimal performance under these constraints. For ToneNet \cite{gao2019tonenet}, the original architecture was designed for larger mel-spectrogram inputs. Since our MFCC inputs have a smaller spatial size, the original architecture led to feature maps that became too small after several convolution and pooling operations, causing runtime errors. To address this, we reduced the number of convolutional blocks from 4 to 3 and replaced the final MaxPooling layers with global average pooling. However, since our focus is on few-shot settings where ToneNet's performance is already limited, this architectural reduction does not affect our comparative conclusions. For CNN-DAE \cite{chen2016tone}, we followed the architecture described in the original paper as faithfully as possible.

\bibliography{custom}

\appendix

\section{Cross-Speaker Generalization}

We further evaluate cross-speaker generalization on the Tone Perfect dataset. Models are fine-tuned on either female (F) or male (M) speakers and tested on female, male, or both genders. Results for both baseline (CNN) models and ToneCL-ZH are reported in Table \ref{tbl:cross_speaker}.

\begin{table*}[h]
\small
\centering
\begin{tabular}{lcc|cccc}
\hline
\textbf{Train} & \textbf{Test} & \textbf{Model} & 1-shot & 3-shot & 5-shot & 10-shot \\
\hline
\multicolumn{7}{c}{\textbf{Female (F)}} \\
\hline
F & F & CNN & 76.7 & 84.5 & 87.3 & 92.3 \\
F & F & ToneCL-ZH & \textbf{80.7} & \textbf{89.5} & \textbf{92.4} & \textbf{96.3} \\
F & M & CNN & 58.5 & 55.6 & 58.1 & 56.8 \\
F & M & ToneCL-ZH & \textbf{65.5} & \textbf{62.1} & \textbf{68.1} & \textbf{66.8} \\
F & Both & CNN & 63.4 & 65.6 & 71.3 & 70.3 \\
F & Both & ToneCL-ZH & \textbf{69.4} & \textbf{72.6} & \textbf{77.3} & \textbf{78.0} \\
\hline
\multicolumn{7}{c}{\textbf{Male (M)}} \\
\hline
M & M & CNN & 53.5 & 66.5 & 82.3 & 83.6 \\
M & M & ToneCL-ZH & \textbf{58.5} & \textbf{74.5} & \textbf{85.3} & \textbf{87.6} \\
M & F & CNN & 50.4 & 61.6 & 58.7 & 51.4 \\
M & F & ToneCL-ZH & \textbf{55.4} & \textbf{66.6} & \textbf{63.7} & \textbf{56.4} \\
M & Both & CNN & 46.5 & 62.6 & 65.9 & 63.0 \\
M & Both & ToneCL-ZH & \textbf{51.5} & \textbf{68.6} & \textbf{71.9} & \textbf{69.0} \\
\hline
\end{tabular}
\caption{Cross-speaker generalization accuracy (\%) for CNN and ToneCL-ZH models.}
\label{tbl:cross_speaker}
\end{table*}

We have several findings. First, same-gender performance is strong for both models, but female speakers consistently outperform male speakers (e.g., F→F: 96.3\% vs. M→M: 87.6\% at 10 shots with ToneCL-ZH). Second, cross-gender transfer (F→M and M→F) leads to a substantial drop in accuracy, highlighting the difficulty of generalizing across genders. Third, ToneCL-ZH outperforms CNN in every setting, with particularly large gains in cross-gender and mixed-gender scenarios (e.g., F→M at 10 shots: 66.8\% vs. 56.8\%). These findings suggest a practical strategy for low-resource language documentation: collecting training data from speakers of the same gender as the target informant can help avoid the performance degradation caused by cross-gender transfer, especially when labeled data is extremely scarce.

\section{Confusion Matrices}

\begin{figure*}
\centering
\includegraphics[width=0.7 \textwidth]{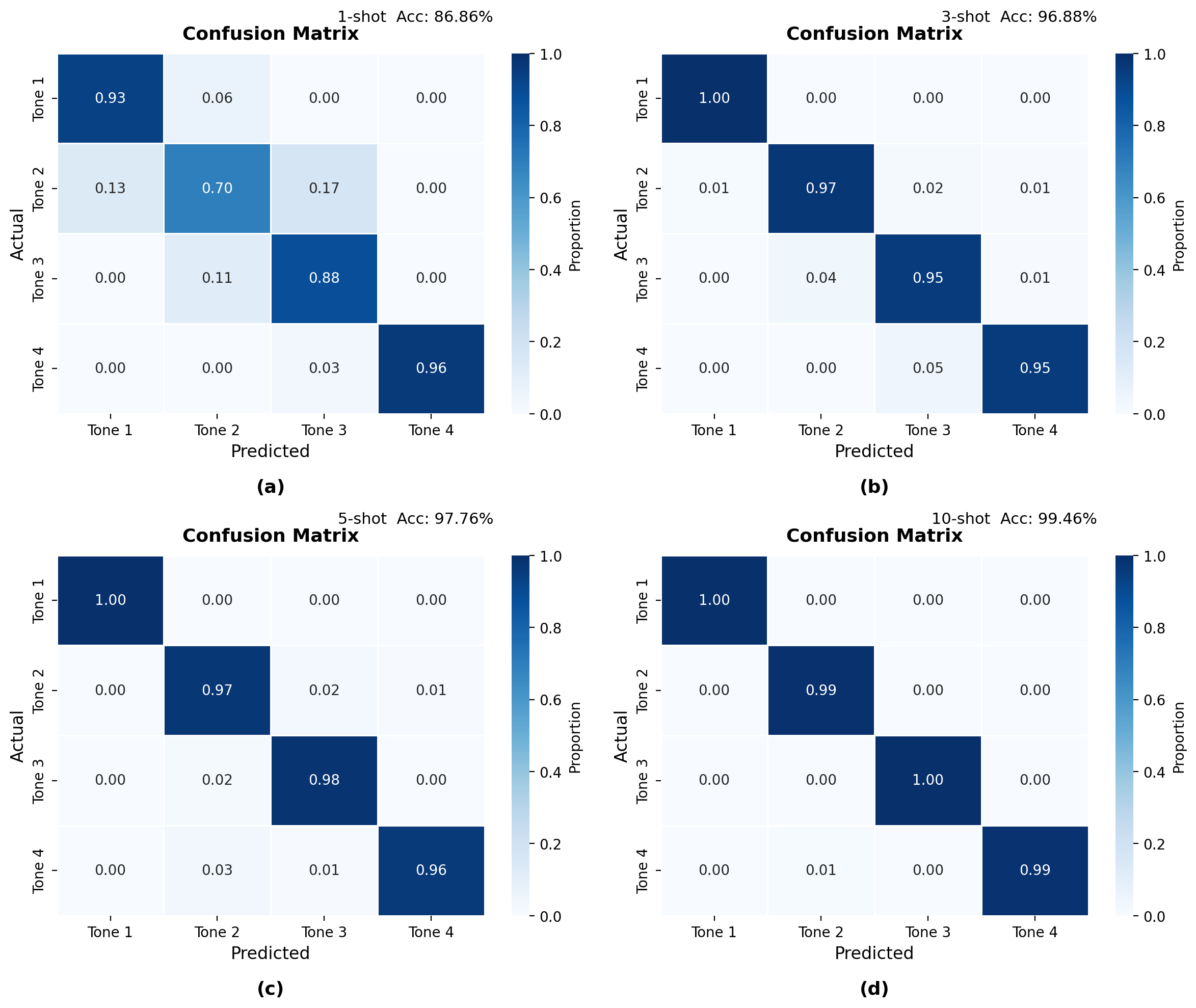}
\caption{Confusion matrices of ToneCL on Tone Perfect (one speaker) with (a) 1 shot, (b) 3 shots, (c) 5 shots, and (d) 10 shots (single seed).}
\label{fig:one_speaker}
\end{figure*}

\begin{figure*}
\centering
\includegraphics[width=0.7 \textwidth]{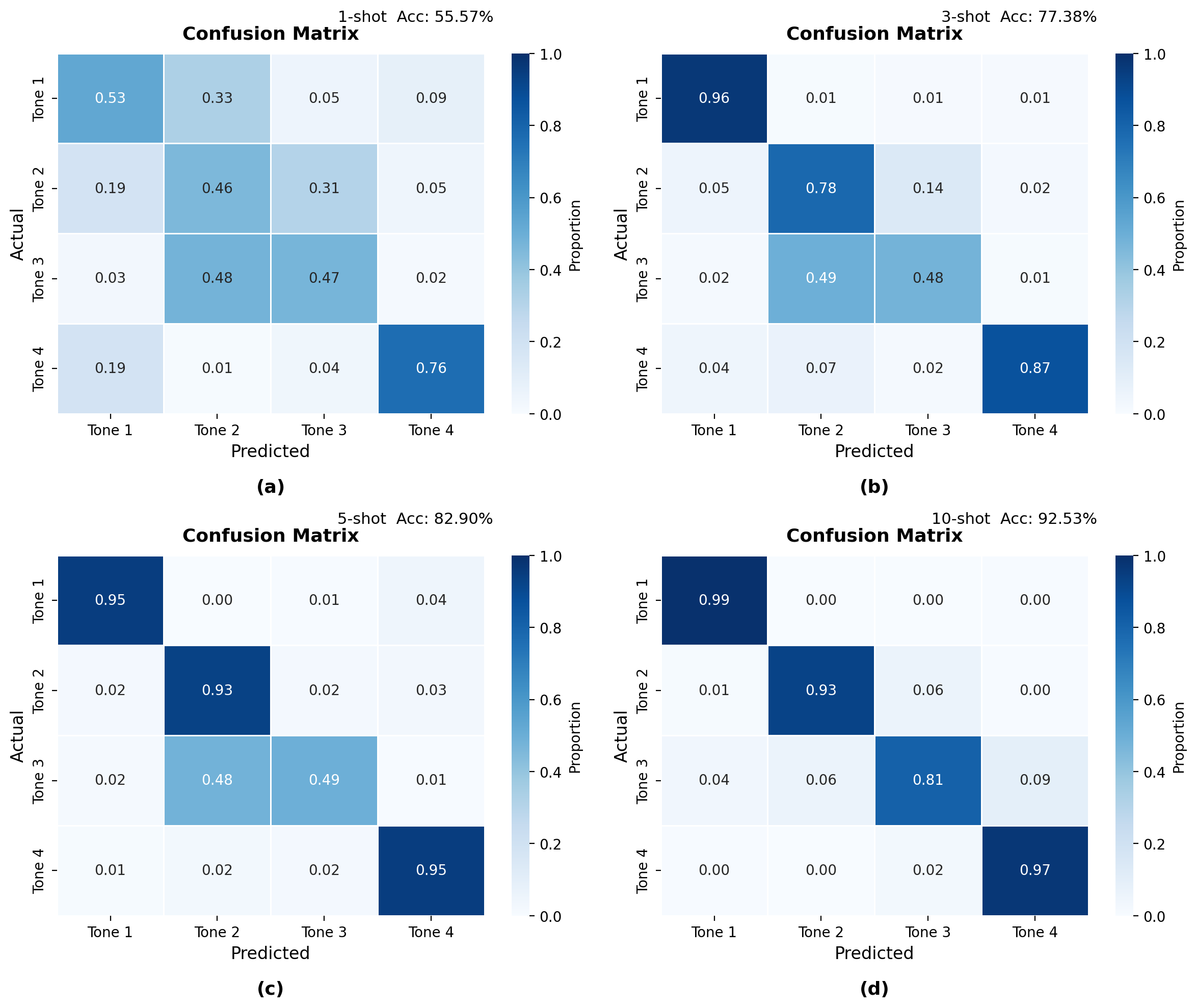}
\caption{Confusion matrices of ToneCL on Tone Perfect (six speakers) with (a) 1 shot, (b) 3 shots, (c) 5 shots, and (d) 10 shots (single seed).}
\label{fig:six_speaker}
\end{figure*}

\begin{figure*}
\centering
\includegraphics[width=0.7 \textwidth]{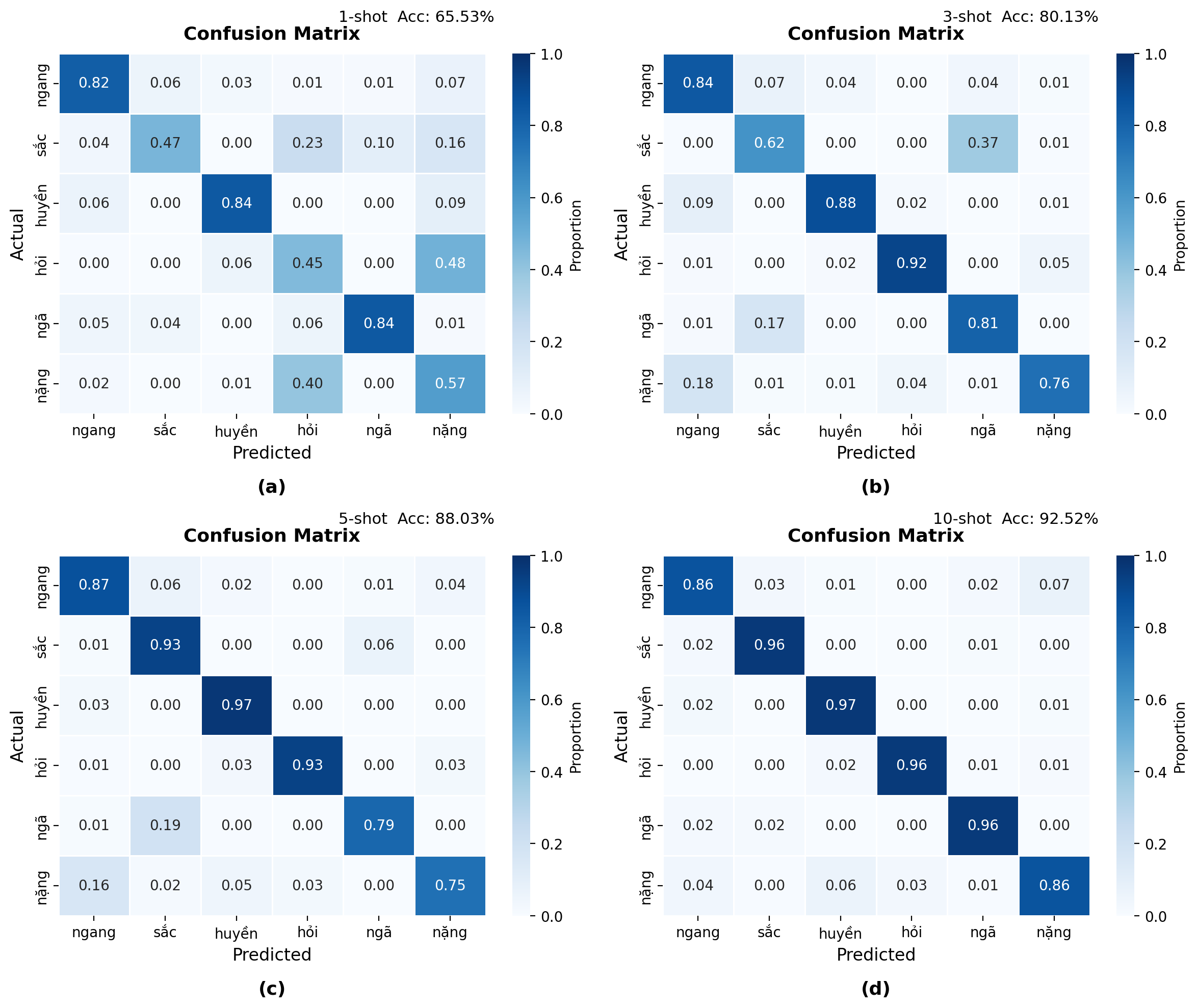}
\caption{Confusion matrices of ToneCL on Google TTS Vietnamese (one speaker) with (a) 1 shot, (b) 3 shots, (c) 5 shots, and (d) 10 shots (single seed).}
\label{fig:vietnamese}
\end{figure*}

Figures \ref{fig:one_speaker}, \ref{fig:six_speaker}, and \ref{fig:vietnamese} present confusion matrices for ToneCL fine-tuned on the one-speaker Tone Perfect, six-speaker Tone Perfect, and Google TTS Vietnamese datasets, respectively, with 1, 3, 5, and 10 shots per tone. These results are from a single representative random seed.

\end{document}